\documentclass[letterpaper, 10pt, conference]{ieeeconf}

\IEEEoverridecommandlockouts
\usepackage{graphicx}
\usepackage{times,cite,flushend}

\usepackage{amsmath,amssymb}
\usepackage{booktabs,array}
\usepackage{multirow}
\usepackage{needspace}
\usepackage{xcolor}

\title{\LARGE \bf
From Wearable Interfaces to Dexterous Policies:\\
Contact Shifts and Tactile Representations}

\author{Ruitong Tian$^{1}$, Fang Xu$^{1}$, Noah B. Wilson$^{2}$, Xianyao Li$^{1}$, and Eric Jing Du$^{1,\dagger}$%
\thanks{$^{1}$Ruitong Tian, Fang Xu, Xianyao Li, and Eric Jing Du are with the Department of Civil and Coastal Engineering, University of Florida, Gainesville, FL, USA.}%
\thanks{$^{2}$Noah B. Wilson is with the Department of Mechanical and Aerospace Engineering, University of Florida, Gainesville, FL 32611, USA.}%
\thanks{$^{\dagger}$Corresponding author: Eric Jing Du (\texttt{eric.du@essie.ufl.edu}). First author: Ruitong Tian (\texttt{ruitongtian@ufl.edu}).}%
}

\begin{document}

\maketitle
\thispagestyle{empty}
\pagestyle{empty}
\raggedbottom

\begin{abstract}
Unlike conventional teleoperation, wearable interfaces allow operators to
collect dexterous demonstrations through their own hand motions while directly
interacting with task objects. This direct interaction reduces dependence on
the target robot during collection, but it also makes the collection hardware
part of the physical process that generates each demonstration. Interface
geometry can influence both how a task is performed and what tactile
observations are recorded for learning. We study two versions of a
DexUMI-family exoskeleton that share the same robot command definition, mapping
procedure, and tactile module type but differ in hand-side geometry. The
revised interface reduces reported physical demand, improves selected device
ratings, enables tactile access in a precision grasp that is mechanically
blocked by the baseline, and produces task-dependent changes in recorded
contact. Lid twisting primarily exhibits a change in contact location, whereas
egg carton opening primarily exhibits a change in contact occurrence. We then
train matched policies using either a binary aggregate input or a
spatial-plus-force input. On lid twisting and egg carton opening, policies trained on
revised-interface demonstrations achieve higher success than those trained on
baseline demonstrations, with a significant pooled interface effect. Across these and two additional tasks,
USB insertion and soldering tool pick and place, the spatial-plus-force input
likewise outperforms binary aggregation. These results show that wearable collection
hardware is part of the data-generation process for robot learning. Such interfaces should therefore be evaluated not only through operator
experience, but also through the tactile interactions they make recordable and
the policy performance their demonstrations support.
\end{abstract}

\section{Introduction}
\label{sec:intro}

\begin{figure*}[t]
\centering
\includegraphics[width=\textwidth]{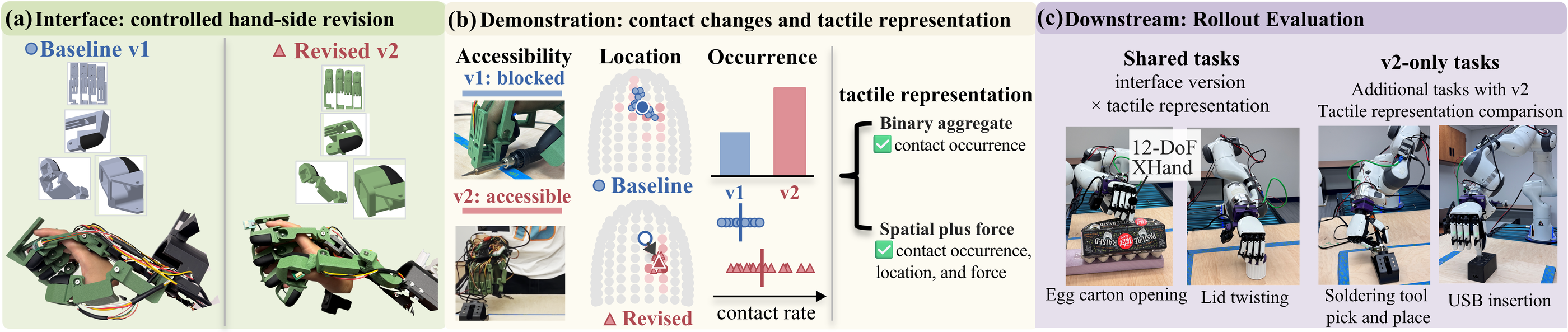}
\caption{A controlled hand-side interface revision changes the physical
interaction between the operator and task object while the robot command,
calibration, and tactile-sensing pipeline remain fixed. The revision produces
task-dependent changes in tactile-pad accessibility, contact location, and
contact occurrence. Binary aggregate and spatial-plus-force inputs preserve
different tactile information, and matched robot rollouts test whether the
retained information affects downstream policy performance.}
\label{fig:overview}
\end{figure*}

Learning dexterous manipulation policies requires demonstrations that preserve
both high-dimensional hand motion and contact-rich interaction. Human-hand
videos and motion-capture systems can collect data without a robot, but
recovering precise contact and robot-feasible joint commands remains
difficult~\cite{wangDexCapScalablePortable2024}. Robot-side teleoperation and
hand-over-hand interfaces provide actions closer to the deployment embodiment,
yet they require the physical robot hand during collection and can be
cumbersome for contact-rich tasks~\cite{liHowTrainRobots2025}. Wearable
interfaces offer a complementary solution: the operator manipulates objects
directly, receives natural contact feedback, and records demonstrations without
bringing the complete robot system into the collection
environment~\cite{xuDexUMIUsingHuman2025,fangDEXOPDeviceRobotic2025,
zhuDexEXOWearabilityFirstDexterous2026,xuRealDexUMIWearableUniversal2026}.

To make these demonstrations transferable, a robot-specific exoskeleton
constrains human motion and measures joint states corresponding to the target
hand. This correspondence is a central strength of systems such as
DexUMI~\cite{xuDexUMIUsingHuman2025}, but it also creates a human--robot design
tension. In our system, the wearable interface maps the operator's fingers to
the joint-command space of a fully actuated 12-DoF XHand. Preserving this
correspondence can impose robot-derived finger spacing, joint clearances, and
linkage geometry on the human hand. Such constraints are particularly
consequential for the thumb, whose natural opposition and circumduction do not
necessarily align with the target robot's kinematic structure. The resulting
mismatch can increase physical burden, restrict motion, and create localized
discomfort during repeated data collection~\cite{mile2025isomorphic,
zhuWHEDWearableHand2026,sunerplacerdaUserRequirementsInvolvement2025,
zhangHumanExoskeletonKinematic2025}.

Unlike a passive sensing system, a wearable collection interface participates
directly in the physical interaction that generates each demonstration. A
restrictive linkage can change a grasp trajectory, while material surrounding
a fingertip sensor can prevent the tactile surface from reaching the object.
The resulting difference may appear as a change in whether contact is
accessible, where contact occurs on the pad, or how often contact is detected.
Tactile representation further determines which of these differences a learned
policy can receive: a binary aggregate input preserves per-finger contact
occurrence while discarding within-pad contact location and force, whereas a
spatial-plus-force input retains contact location on the pad and aggregate
fingertip force~\cite{helmutTactileConditionedDiffusionPolicy2025,
xueReactiveDiffusionPolicy2025,yuanRobotSynesthesia2024,
huang3DViTacLearningFineGrained2025,huangSpatiallyAnchoredTactile2025,
krohnMultiResolutionTactile2026}.

This motivates our central question: when a revision of wearable collection
hardware changes physical contact during demonstration collection, how do the
collection interface and the tactile representation each affect downstream
robot policy performance? We study this question through a controlled hand-side revision
within a robot-aligned wearable interface. The baseline and revised interfaces
retain the same 12-DoF XHand command definition, encoder assignment,
calibration procedure, tactile module type, and robot platform, while differing
in hand-side geometry. We first test whether the revision changes operator
experience and tactile-pad accessibility. We then measure whether it changes
where or how often contact is recorded in matched demonstrations. Finally, we train policies that cross the two collection interfaces with a
binary aggregate or a spatial-plus-force input and evaluate them on the same
XHand, which measures the contribution of each factor to policy success.

Our contributions are:
\begin{itemize}
    \item We show that wearable interface geometry can alter the tactile data generated during demonstration collection while the robot
    command, mapping, calibration, and sensing pipeline remain fixed.
    \item We characterize these effects through task-dependent changes in
    tactile-pad accessibility, within-pad contact location, and contact
    occurrence across matched demonstrations.
    \item Through matched real-robot rollouts that cross collection interface with
    tactile representation, we show that revised-interface demonstrations and a
    spatial-plus-force input each increase dexterous policy success in the pooled
    comparisons.
\end{itemize}

\section{Related Work}
\label{sec:related}

\subsection{Demonstration Collection Interfaces}
Dexterous demonstrations can be collected from human observations, motion
capture, and robot-mediated interfaces. Human video scales to diverse behavior,
and motion-capture systems such as DexCap improve wrist and finger tracking
without the target robot, but both still require human motion and contact to be
converted into robot-feasible actions~\cite{wangDexCapScalablePortable2024}.
Robot-mediated teleoperation and hand-over-hand collection provide commands
closer to the deployment platform but retain dependence on robot hardware
during collection~\cite{liHowTrainRobots2025}. Portable handheld and
visuotactile systems reduce that dependence and support direct object
interaction through gripper-like or task-specific collection
embodiments~\cite{chiUniversalManipulationInterface2024,
liuViTaMInLearningContactRich2025,wuFreeTacManRobotfree2026,
zhuTouchWildLearning2025,xuExUMIExtensible2025}.

Wearable interfaces maintain direct correspondence with a target hand while
allowing the operator to manipulate objects through their own fingers.
Representative systems use robot-specific kinematics, shared mechanisms, or
wearable sensing and feedback to acquire transferable multifinger
demonstrations~\cite{xuDexUMIUsingHuman2025,
zhuDexEXOWearabilityFirstDexterous2026,fangDEXOPDeviceRobotic2025,
xuRealDexUMIWearableUniversal2026,osmo2025tactileglove,
mile2025isomorphic}. This correspondence also places the collection hardware
inside the physical interaction. Wearability-oriented systems such as DexEXO and WHED introduce new exoskeleton
designs for comfortable, natural collection and evaluate each design as a
complete system~\cite{zhuDexEXOWearabilityFirstDexterous2026,
zhuWHEDWearableHand2026}. We instead hold the robot command, calibration, and
sensing fixed and compare the tactile data and policies produced before and
after a controlled hand-side revision.

\subsection{Mechanical Correspondence and Human Interaction}
Mechanical correspondence makes wearable demonstrations transferable, but it
also allows robot-hand geometry to constrain the operator. Finger spacing,
joint axes, clearances, and thumb geometry designed around a robot's
configuration can restrict natural finger trajectories or create localized
pressure. Prior exoskeleton systems therefore emphasize passive accommodation,
wearability, calibration, or returned haptic and fingertip
cues~\cite{chaoExoViHaCrossPlatformExoskeleton2025,
chenDexViTacCollectingHuman2026a,zhangDOGloveDexterousManipulation2025,
zhuWHEDWearableHand2026,sunerplacerdaUserRequirementsInvolvement2025,
zhangHumanExoskeletonKinematic2025}.

For demonstration learning, these constraints can also change the data itself.
A restricted finger trajectory can alter a grasp, while material surrounding a
fingertip sensor can prevent a physically possible contact from being measured.
We therefore treat operator experience and recorded contact as two consequences
of the same hand-side interface revision.

\subsection{Tactile Representation and Downstream Learning}
Prior tactile policies use binary contact signals, per-finger force summaries,
dense taxel arrays, spatial features, and multiresolution tactile
representations~\cite{helmutTactileConditionedDiffusionPolicy2025,
xueReactiveDiffusionPolicy2025,yuMimicTouchLeveragingMultimodal2025,
yuanRobotSynesthesia2024,huang3DViTacLearningFineGrained2025,
huangSpatiallyAnchoredTactile2025,krohnMultiResolutionTactile2026,
xuETacLightweightEfficient2026}. These inputs preserve different properties of
physical interaction. A binary aggregate signal can retain whether a finger is
in contact while discarding within-pad location and force distribution,
whereas a spatial tactile representation can retain those distinctions. A
contact difference can therefore be present in the recorded tactile array yet
disappear before reaching the policy. We compare two complete policy inputs---binary aggregate and spatial plus
force---derived from the same tactile measurements and evaluated on the same
robot.

\section{Interface Revision and Policy Inputs}
\label{sec:hardware}

We use the hand-side revision as a controlled study mechanism for changing
wearable geometry while keeping the robot command and sensing pipeline fixed.
The baseline interface (v1) and the revised interface (v2) share the same
12-DoF XHand control and tactile-sensing pipeline while differing in finger clearance, fingertip housing, and thumb
seat and link geometry.

\begin{figure}[t]
\centering
\includegraphics[width=\columnwidth]{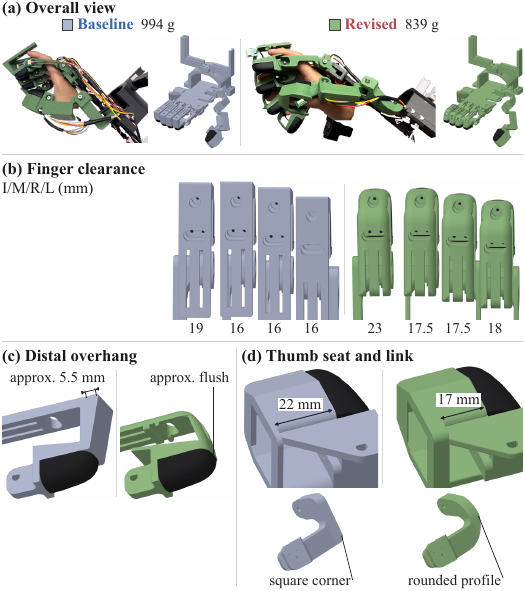}
\caption{Hand-side interface revision. The baseline and revised interfaces
retain the same tactile modules and robot mapping but differ in per-finger
clearance, fingertip housing geometry, and thumb seat and link geometry. The
baseline fingertip housing extends approximately 5.5\,mm beyond the tactile
pad, whereas the revised housing is approximately flush with the pad.}
\label{fig:hardware}
\end{figure}

\subsection{Hand-Side Interface Revision}
\label{sec:hardware-a}

Both interfaces are right-hand exoskeletons that map the operator's finger
motion to a fully actuated 12-DoF XHand. Each carries a wrist RGB camera, an
iPhone pose tracker, and three fingertip tactile modules on the thumb, index,
and middle fingers---the three fingers that contact the objects in the evaluated
grasps, and the fingers instrumented in the released design---with 120 taxels per
module, three-axis force sensing, and
an acquisition rate of approximately 59\,Hz. The baseline interface follows
the released DexUMI design. The revised interface changes the hand-side
geometry while retaining the same sensing hardware, encoder assignment, and
robot platform.

\textbf{Finger clearance.} The measured clearances for the index, middle,
ring, and little fingers are 19, 16, 16, and 16\,mm in the baseline interface
and 23, 17.5, 17.5, and 18\,mm in the revised interface, respectively. The
revision increases the available space for all four fingers while retaining
the same distal tactile-module positions.

\textbf{Fingertip housing.} The baseline housing extends approximately
5.5\,mm beyond the tactile pad. The revised housing follows the pad outline,
bringing the rounded hood approximately flush with the tactile surface
(Fig.~\ref{fig:hardware}c). This change removes surrounding material that
previously blocked the tactile surface from reaching some objects. The tactile
modules remain unchanged.

\textbf{Thumb seat and link profile.} The baseline thumb-tip holder provides
22\,mm of seat clearance, whereas the revised holder provides 17\,mm. The
reduced seat clearance is intended to limit free relative motion at the
fingertip. Separately, the baseline link includes a square corner between the
first and second joints, whereas the revised link uses a wrapped, rounded
profile intended to provide more external space for thumb opposition and
circumduction.

We evaluate these changes as one integrated hand-side revision and do not
attribute downstream observations to any single geometric modification.

\subsection{Shared XHand Command and Collection Pipeline}
\label{sec:hardware-b}

Both interfaces use the same 12-DoF XHand command definition,
encoder-channel assignment, mapping code, five-keypose calibration procedure,
control frequency, tactile module type, and robot platform
(Table~\ref{tab:fixed}). The released DexUMI pipeline replays recorded actions
on the physical XHand and applies segmentation, inpainting, and robot-hand
compositing~\cite{xuDexUMIUsingHuman2025}. Visual processing is identical across all policy comparisons, so the experimental factors remain
interface version and tactile representation.

\begin{table}[t]
\makeatletter
\long\def\@makecaption#1#2{%
  \noindent{\footnotesize #1: #2\par}\vspace{4pt}}
\makeatother
\centering
\caption{\textbf{Hardware revision and fixed pipeline.} Changed hand-side geometry and shared components.}
\label{tab:fixed}
\footnotesize
\setlength{\tabcolsep}{2pt}
\renewcommand{\arraystretch}{1.02}
\begin{tabular}{@{}p{0.42\columnwidth}p{0.23\columnwidth}p{0.315\columnwidth}@{}}
\toprule[1.5pt]
Component & Baseline (v1) & Revised (v2) \\
\midrule
Finger clearance, I/M/R/L (mm) & 19/16/16/16 & 23/17.5/17.5/18 \\
Thumb seat clearance (mm) & 22 & 17 \\
Distal overhang (mm) & approx.\ 5.5 & approx.\ flush \\
Thumb link & square corner & rounded profile \\
Hand-mounted mass (g) & 994 & 839 \\
\bottomrule[1.5pt]
\end{tabular}
\par\vspace{2pt}
\parbox{0.96\columnwidth}{\scriptsize Held fixed: XHand
12-DoF command definition; encoder parts, channel assignment, mapping code and
five-keypose calibration; three fingertip modules of 120 taxels $\times$ 3
force axes.}
\end{table}

\subsection{Recorded Data and Tactile Policy Inputs}
\label{sec:record}

Each demonstration records raw wrist-camera frames, wrist pose, encoder angles, and a
$3{\times}120{\times}3$ tactile array from the thumb, index, and middle
fingertips, where each of the 120 taxels measures three-axis force. From the
same stored tactile array, we construct two policy inputs that differ in the
contact information they retain.

The XHand is equipped with magnet-based Hall-effect tactile arrays, each
providing three-axis readings at 120 taxels. The exoskeleton carries the same
module type on each finger, and both sides are read with the same taxel indexing.

Let $\mathbf F_{tkj}\in\mathbb R^{3}$ denote the three-axis force reading at
frame $t$ from taxel $j$ on finger $k\in\{\text{thumb},\text{index},\text{middle}\}$.
After subtracting the reference
$\mathbf F^{\mathrm{ref}}_{kj}$, the binary aggregate input follows the released
DexUMI baseline:
\begin{equation}
\label{eq:flag}
\begin{aligned}
\tilde{\mathbf f}_{tkj}&=\mathbf F_{tkj}-\mathbf F^{\mathrm{ref}}_{kj},\\
b_{tk}&=\mathbf{1}\!\left[\max_j\big\lVert\tilde{\mathbf f}_{tkj}\big\rVert_2\ge\tau\right],
\qquad \tau=10.
\end{aligned}
\end{equation}
Here, $\tau$ is expressed in sensor units and equals the released DexUMI
threshold for the XHand~\cite{xuDexUMIUsingHuman2025}. The reference is the first frame of
each recorded demonstration during dataset construction and the mean of ten samples
collected after hand reset during rollout. This representation yields one
contact flag per instrumented finger, preserving contact occurrence while discarding within-pad contact location and
force.

The \emph{spatial-plus-force input} replaces the per-finger flag with a
360-taxel binary contact map, thresholded as in (\ref{eq:flag}), and appends one
summed three-axis force vector per fingertip:
\begin{equation}
\label{eq:spatial}
\begin{aligned}
\mathbf s_t=\operatorname{concat}\!\Big(&\big\{\mathbf{1}\big[\lVert\tilde{\mathbf f}_{tkj}\rVert_2\ge\tau\big]\big\}_{k,j},\\
&\big\{\textstyle\sum_{j=1}^{120}\tilde{\mathbf f}_{tkj}\big\}_{k}\Big)\in\mathbb R^{369}.
\end{aligned}
\end{equation}
It therefore retains contact occurrence, within-pad contact location, and
aggregate three-axis fingertip force. Both inputs use the references defined for
(\ref{eq:flag}) during dataset construction and rollout. The comparison between
these two complete inputs is the representation-level test used in the robot
evaluation.

\section{Evaluation Protocol}
\label{sec:measure}

The participant study asks whether the hand-side revision changes operator burden and device
experience. The accessibility test asks whether the revision changes which
fingertip contacts can be physically recorded. The shared-task demonstration
analysis asks whether the revision changes contact location or contact
occurrence. A dataset-separability diagnostic tests which tactile feature
family most clearly distinguishes the two interface datasets. Finally, matched robot rollouts measure how the collection interface and the
tactile representation each affect policy success.

\subsection{Operator Experience and Tactile Accessibility}
\label{sec:measure-user}

Twenty (N=20) right-handed participants evaluated both interfaces in a randomized
within-subject study. The
protocol was approved by the institutional review board, and all participants
gave informed consent (IRB number hidden for anonymous review). Participants were assigned to a baseline-first or
revision-first order, with version labels and design intent concealed. With
each interface, they completed lid twisting and egg carton opening before
providing workload and device-experience ratings.

Workload was assessed using the six raw NASA-TLX dimensions on a 0--20
scale~\cite{hartNasaTaskLoadIndex2006}, with Performance anchored so that higher
values indicate better self-rated performance; overall workload is the
unweighted mean of the six dimensions with Performance reversed. Device experience was assessed using
six custom items with the five-point response format used by
SUS~\cite{brookeSUSQuickDirty1996}. Physical Demand was the prespecified
primary outcome. Paired differences were tested with two-sided Wilcoxon
signed-rank tests, with paired sign-flip tests as a sensitivity analysis, and
we report standardized paired differences $d_z$. The remaining NASA-TLX
outcomes and the device-specific items were corrected within their respective
families using false-discovery-rate (FDR) control.

As a task-specific accessibility check, we used the precision grasp on the
soldering-tool handle shown in Fig.~\ref{fig:overview}b. This test asks whether
the revised fingertip housing changes the physical ability of the tactile pad
to reach the object. Pad access was available when the tactile surface could
reach the handle before the surrounding fingertip housing.

\subsection{Demonstration Collection and Contact Measures}
\label{sec:measure-a}

\begin{figure}[t]
\centering
\includegraphics[width=\columnwidth]{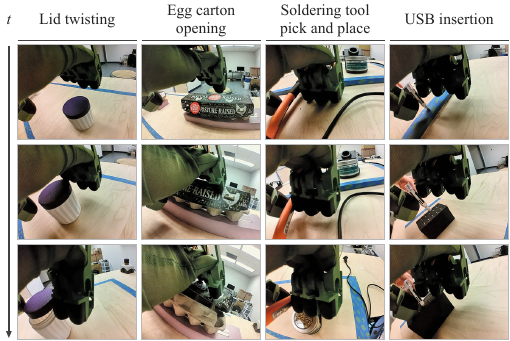}
\caption{Wearable demonstration collection with the revised interface.
Wrist-camera views from one demonstration per task; rows run forward in time:
before measured fingertip contact, early in contact, and late in contact.}
\label{fig:datacollection}
\end{figure}

We collected lid-twisting and egg-carton-opening demonstrations from human participants with both
interfaces for the contact analysis and policy training. A session is a consecutive block of demonstrations recorded in one sitting. Demonstrations with the two interfaces
were collected alternately rather than in two separate phases. The contact analysis is computed at
the session level over 20 sessions per interface for lid twisting and 16 baseline
and 17 revised sessions for egg carton opening; baseline egg-carton sessions
recorded before an encoder repair were excluded before analysis. Soldering tool pick and place (311 demonstrations) and USB insertion (319
demonstrations) were collected only with the revised interface and enter the robot evaluation
(Fig.~\ref{fig:datacollection}). Both tasks require grasping thin objects: with the
baseline fingertip housing, which extends about 5.5\,mm beyond the tactile pad,
contact falls on the housing before the pad, so baseline demonstrations would not
provide valid tactile data for the representation comparison.

The collection session is the unit of inference. Demonstration quantities are averaged within session, differences between collection conditions are tested
with session-level permutation tests (20 versus 20 sessions on lid, 16 versus
17 on egg), and standardized differences are computed on session means.
An automated check screened every saved demonstration for encoder-reference drift,
encoder or tactile dropouts, flat channels, and insufficient camera overlap.

Two contact statistics were fixed before analysis. With $\tilde f^{z}_{tkj}$ the normal component of $\tilde{\mathbf f}_{tkj}$
and $\tau$ as in (\ref{eq:flag}),
\begin{equation}
\label{eq:contact}
\begin{aligned}
c_{tkj}&=\mathbf{1}\big[\tilde f^{z}_{tkj}\ge\tau\big],\qquad
O_{kj}=\tfrac{1}{T}\textstyle\sum_{t}c_{tkj},\\
r_{k}&=\tfrac{1}{T}\textstyle\sum_{t}\mathbf{1}\big[\textstyle\sum_{j}c_{tkj}>0\big].
\end{aligned}
\end{equation}
The rate $r_{k}$ thresholds the normal force of each taxel, whereas the policy
flag $b_{tk}$ in (\ref{eq:flag}) thresholds the largest three-axis taxel force
magnitude: the policy input keeps the released DexUMI thresholding rule
unchanged, while
the contact statistics use the normal component that defines the occupancy map
in (\ref{eq:contact}). Both use the same threshold $\tau$.
The reported contact rate and the binary policy input are therefore different
statistics at the same threshold. \emph{Contact location} is described by the
occupancy map $O_{kj}$, its centroid in pad coordinates (mm), and the
displacement of that centroid between collection conditions,
\begin{equation}
\label{eq:centroid}
\boldsymbol{\mu}_{k}=\frac{\sum_{j}O_{kj}\,\boldsymbol{x}_{j}}{\sum_{j}O_{kj}},\qquad
\delta_{k}=\big\lVert\bar{\boldsymbol{\mu}}^{\mathrm{rev}}_{k}
-\bar{\boldsymbol{\mu}}^{\mathrm{base}}_{k}\big\rVert_{2},
\end{equation}
where $\boldsymbol{x}_{j}\in\mathbb{R}^{2}$ is the taxel position in the
pad plane (mm), $\boldsymbol{\mu}_{k}$ is computed per demonstration and
$\bar{\boldsymbol{\mu}}^{\mathrm{base}}_{k}$ and $\bar{\boldsymbol{\mu}}^{\mathrm{rev}}_{k}$ are the
means of session means for the baseline and revised interfaces. \emph{Contact occurrence} is the per-finger contact rate $r_{k}$,
together with the number of active taxels during detected contact when needed
to distinguish contact frequency from contact-patch size. Occupancy and rate
are normalized by demonstration length. Both statistics come from the tactile pads
and do not depend on encoder scale.

\subsection{Dataset Separability Diagnostic}
\label{sec:measure-c}

As a supporting diagnostic, a linear classifier predicts interface version
from demonstration-level tactile summaries while holding out complete collection
sessions. We compare contact-location features with contact-amount-and-timing
features. The area under the ROC curve (AUROC) measures how clearly each
feature family separates the two datasets, with shuffled interface labels
providing a chance control. These diagnostic features are distinct from the
policy inputs.

\subsection{Policy Evaluation}
\label{sec:measure-d}

On each shared task we use Diffusion
Policy~\cite{chiDiffusionPolicyVisuomotor2023} to train four policies by
crossing collection interface with the binary aggregate and spatial-plus-force
inputs. All policies share the released DexUMI
configuration~\cite{xuDexUMIUsingHuman2025}: a DINO ViT-S/8 image encoder,
fine-tuned with the policy, embeds one $224{\times}224$ wrist-camera image; the
tactile vector is concatenated with this embedding as the global condition of a
1-D temporal U-Net, which predicts 16 future actions (wrist pose and 12
hand-motor commands) from a single observation frame. No proprioceptive input is
used. Each policy is trained for 400 epochs with an effective batch size of 297.
USB insertion and soldering tool pick and place use revised-interface
demonstrations under both inputs, giving twelve policy--task cells in all.
Each shared-task condition trains on a matched budget of
341 lid-twisting and 252 egg-carton demonstrations; after the tactile quality screen,
the larger dataset of each task is randomly subsampled to this budget. All policies deploy on the same
12-DoF XHand mounted on a Franka Emika robot arm. The collection interface enters only through the training data. Each policy--task cell is evaluated in 20 rollouts over the same pre-generated
task-specific initial-state list. A rollout is successful under the following
criteria:
\begin{itemize}
    \item \emph{Lid twisting}: the hand aligns with the lid, twists it once, and
    lifts it from the jar.
    \item \emph{Egg carton opening}: the hand aligns with the carton and opens it.
    \item \emph{Soldering tool pick and place}: the hand picks up the soldering tool
    and places it in its holder, including placements that remain unsteady.
    \item \emph{USB insertion}: the hand picks up the USB stick and inserts it into
    the hub, including loose insertions.
\end{itemize}

For each cell we report the success rate over 20 rollouts. The representation contrast compares the two inputs within each task
and collection interface (six comparisons), and the interface contrast compares
revised with baseline demonstrations within each shared task and input (four
comparisons). Each contrast is summarized by a Mantel--Haenszel common odds ratio
across its comparisons and tested with an exact conditional test stratified by
comparison, treating rollouts as independent trials.

\section{Results}
\label{sec:results}

\begin{table*}[!t]
\makeatletter
\long\def\@makecaption#1#2{%
  \noindent{\footnotesize #1: #2\par}\vspace{4pt}}
\makeatother
\centering
\caption{\textbf{Policy evaluation.} We report success rates over 20
rollouts per cell.
The shared-task experiments compare combinations of collection interface
(baseline or revised) and tactile representation (binary aggregate or spatial
plus force); USB insertion and soldering tool pick and place compare the two
tactile representations using revised-interface demonstrations.}
\label{tab:rollouts}
\small
\setlength{\tabcolsep}{3pt}
\renewcommand{\arraystretch}{1.02}
\begin{tabular}{@{}l>{\centering\arraybackslash}p{1.3cm}>{\centering\arraybackslash}p{1.3cm}@{\hspace{14pt}}>{\centering\arraybackslash}p{1.3cm}>{\centering\arraybackslash}p{1.3cm}@{\hspace{14pt}}>{\centering\arraybackslash}p{2.2cm}>{\centering\arraybackslash}p{2.2cm}@{}}
\toprule[1.5pt]
& \multicolumn{4}{c@{\hspace{14pt}}}{Shared tasks: interface $\times$ tactile input}
& \multicolumn{2}{c@{}}{Revision-only tasks: tactile input} \\
\cmidrule(r){2-5}\cmidrule(l){6-7}
& \multicolumn{2}{c@{\hspace{14pt}}}{Lid twisting}
& \multicolumn{2}{c@{\hspace{14pt}}}{Egg carton opening}
& Soldering tool pick and place & USB insertion \\
\cmidrule(r){2-3}\cmidrule(r){4-5}\cmidrule(lr){6-6}\cmidrule(l){7-7}
Tactile input & Baseline & Revised & Baseline & Revised & Revised & Revised \\
\midrule
Binary aggregate & 0.15 & 0.40 & 0.40 & 0.60 & 0.35 & 0.20 \\
Spatial plus force & 0.35 & 0.55 & 0.65 & 0.80 & 0.45 & 0.30 \\
\bottomrule[1.5pt]
\end{tabular}
\par\vspace{2pt}
\parbox{0.84\textwidth}{\scriptsize Shared-task policies use matched
demonstration budgets per interface.
All cells use one trained policy and the same 20-entry task-specific
initial-state list.}
\end{table*}

\subsection{Operator Experience and Tactile Accessibility}
\label{sec:results-user}

\begin{figure}[!t]
\centering
\includegraphics[width=\columnwidth]{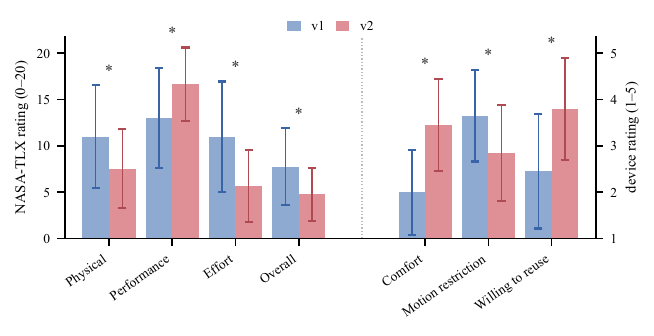}
\caption{The hand-side revision reduces reported physical demand and improves selected
device ratings. Paired ratings ($N=20$; mean $\pm$ SD) are shown for
NASA-TLX (left, 0--20) and device items (right, 1--5). Displayed items
pass the primary test or family-specific correction; higher performance,
comfort, and reuse scores are favorable.}
\label{fig:tlx}
\end{figure}

The revised interface reduced the reported physical burden of data collection
(Fig.~\ref{fig:tlx}). Mean Physical Demand decreased from $11.0\pm5.5$ under the baseline interface to
$7.5\pm4.3$ under the revised interface ($d_z=-0.61$, $p=0.017$). Among the
secondary workload outcomes, Effort and overall workload decreased (both
$q=0.009$) and self-rated Performance improved from 13.0 to 16.6 ($q=0.024$),
whereas Mental Demand, Temporal Demand, and Frustration did not differ after
correction. Device ratings showed higher Comfort, increasing from $2.0\pm0.9$ to
$3.5\pm1.0$ ($d_z=1.14$, $q=0.002$), together with lower motion restriction and
greater willingness to reuse the device. These results show that the revision
achieved its operator-side design objective.

Trimming the fingertip housing also restores tactile-pad access in the
evaluated precision grasp of the soldering-tool handle. Under the baseline
geometry, the housing edge reaches the handle before the tactile pad,
preventing direct tactile-pad contact; the flush revised hood allows the pad
to contact the handle directly (Fig.~\ref{fig:hardware}c; Fig.~\ref{fig:overview}b), enabling
the soldering-tool demonstrations used in the robot evaluation.

\subsection{Task-Dependent Contact Changes}
\label{sec:results-record}

The same interface revision produces different tactile consequences across the
shared tasks. Lid twisting primarily changes where thumb contact occurs,
whereas egg carton opening primarily changes how often index contact is
detected.

\begin{figure*}[t]
\centering
\includegraphics[height=128pt]{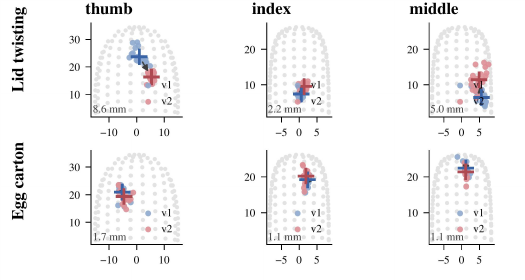}\hspace{18pt}%
\includegraphics[height=128pt]{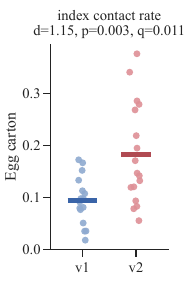}
\caption{Contact location and contact occurrence on the two shared tasks. Left: each
panel shows the fingertip pad, session-mean contact centroids (v1 blue, v2 rose),
interface means as crosses, and their displacement in mm; the thumb during lid
twisting shows the clearest location shift. Right: session-level index contact
frame rate during egg carton opening; bars mark interface means. The analysis
includes 20 sessions per interface for lid twisting and 16 baseline and 17 revised
sessions for carton opening.}
\label{fig:contact}
\end{figure*}

\textbf{Contact location.} During lid twisting, the thumb shows the clearest
change in contact location. The session-mean contact centroid shifts by
8.6\,mm in the pad plane from the baseline to the revised interface, including
7.4\,mm along the pad long axis, with a session-level permutation
$p<0.001$ (Fig.~\ref{fig:contact}, left); the shift remains 8.3 and 8.2\,mm at
$\tau=5$ and $\tau=2$. This establishes a task-specific spatial
change in the recorded tactile interaction.

\textbf{Contact occurrence.} Egg carton opening shows a different pattern.
Index contact frame rate increases from 0.09 under the baseline interface to
0.18 under the revised interface, with an effect size of $d=1.15$ on session
means ($p=0.003$, $q=0.011$; Fig.~\ref{fig:contact}, right). The number of active taxels during detected
contact changes little, from 2.7 to 3.0, indicating that the dominant
difference is how often index contact is detected rather than the size of the
active contact patch.

\textbf{Supporting diagnostic.} The dataset-separability diagnostic supports
the task-dependent interpretation. On lid twisting, contact-location features
separate the interface datasets with AUROC 0.99, compared with 0.84 for
contact-amount-and-timing features. On egg carton opening, the corresponding
AUROCs are 0.82 and 0.80, indicating that both feature families retain
comparable information about the interface difference. Shuffled labels remain
near chance. The diagnostic therefore supports a predominantly spatial
difference on lid twisting and a broader occurrence-related difference on
carton opening.

\begin{figure*}[t]
\centering
\includegraphics[width=\textwidth]{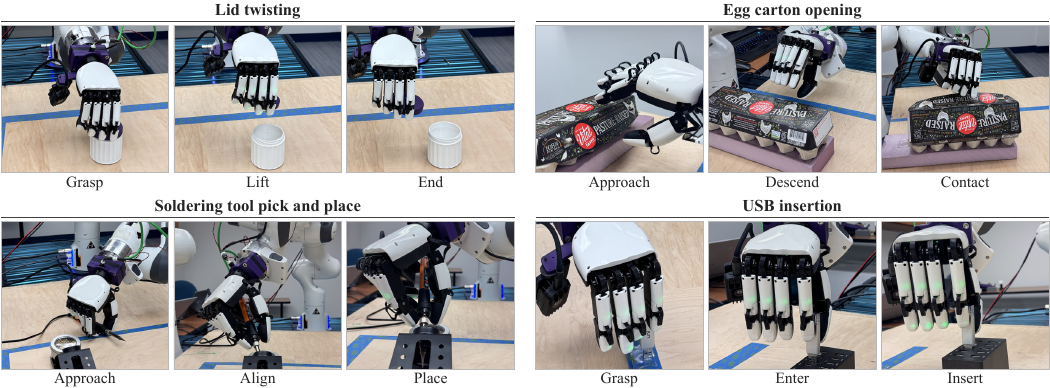}
\caption{Autonomous policy rollouts with the 12-DoF XHand mounted on a Franka
Emika arm: lid twisting, egg carton opening, soldering tool pick and place, and USB
insertion. Frames within each task are ordered in time. The shared tasks cross
collection interface with tactile input; the additional tasks compare the two
inputs using revised-interface demonstrations.}
\label{fig:tasks}
\end{figure*}

\subsection{Policy Performance}
\label{sec:results-robot}

Table~\ref{tab:rollouts} reports the robot evaluation, and
Fig.~\ref{fig:tasks} shows representative rollouts. 
\textbf{Tactile representation.} The spatial-plus-force input achieves a higher
success rate than binary aggregation in all six task--interface comparisons: by
0.20 and 0.15 on lid twisting with baseline and revised demonstrations, by 0.25
and 0.20 on egg carton opening, and by 0.10 on both USB insertion and soldering tool pick and place. Pooled over these comparisons, the success rate rises
significantly from 0.35 to 0.52 (common odds ratio 2.1, $p=0.010$).

\textbf{Collection interface.} Revised-interface demonstrations achieve a higher
success rate than baseline demonstrations in all four shared-task comparisons. On
lid twisting, the success rate increases from 0.15 to 0.40 with the binary
aggregate input and from 0.35 to 0.55 with the spatial-plus-force input; on egg
carton opening, it increases from 0.40 to 0.60 and from 0.65 to 0.80. Pooled, the
success rate rises significantly from 0.39 to 0.59 (common odds ratio 2.5, $p=0.013$). The
interface gains are of similar size under the two inputs (0.20--0.25 and
0.15--0.20), so the results indicate separate contributions of interface and
representation rather than an interface effect that requires the richer input.
Lid twisting and USB insertion remain the most difficult tasks.

\section{Discussion and Limitations}
\label{sec:discussion}

\textbf{Wearable geometry shapes data generation.} Wearable collection
hardware participates directly in the physical interaction that generates a
demonstration. The revised interface reduces operator burden and restores
tactile-pad accessibility in a precision grasp that is blocked by the baseline
geometry. On the shared tasks, it also changes the recorded tactile interaction
in different ways, producing a spatial shift in thumb contact during lid
twisting and a change in index contact occurrence during egg carton opening.
These results show that wearable geometry affects both which interactions can
be collected and what tactile evidence enters the dataset.

\textbf{Representation determines what remains available to the policy.} The binary aggregate input preserves per-finger contact occurrence but discards
within-pad contact location and force, whereas the spatial-plus-force input
retains both. In the matched robot comparison, the spatial-plus-force input
and revised-interface demonstrations each increase success significantly in the
pooled comparisons, and the interface gain is of similar size under both inputs. Retaining the raw tactile arrays during
collection makes this comparison possible without recollecting demonstrations.

\textbf{Remaining hardware constraints.} A more wearable interface does not
remove every constraint of robot-specific collection. Collection itself proceeds
without the robot, but closing the visual gap between exoskeleton and robot-hand
observations still relies on robot-side replay and image compositing, as in
DexUMI~\cite{xuDexUMIUsingHuman2025}. Robot-hand precision also limits fine manipulation. Backlash and friction make the
fingertip position reached for a motor command depend on the direction of motion,
so the regression from exoskeleton encoder values to hand-motor commands is
accurate in only one direction, either closing or opening, and the XHand
command-to-fingertip mapping shifts slightly over time and across
reboots~\cite{xuDexUMIUsingHuman2025}. Even with
improved wearability, the precision of the robot hand and the mechanical
constraints of the exoskeleton still limit which dexterous tasks can be collected
with the interface and reproduced on the robot.

\textbf{Limitations.} The study evaluates an integrated hand-side revision, so
the individual effects of finger clearance, fingertip housing, and thumb
geometry are not isolated. Each condition uses one trained policy instance, so rollout
variability reflects initial-state variation rather than variation across policy
training runs. The comparison between
binary aggregate and spatial plus force is a representation-level comparison
and does not isolate the contribution of spatial structure from force
information within the richer input. The conclusions are limited to the
evaluated objects and tasks.

\section{Conclusion}
\label{sec:conclusion}

This work studies wearable collection hardware as part of the data-generation
process for dexterous robot learning. A controlled hand-side revision reduces
reported operator burden, restores tactile-pad accessibility in a precision
grasp, and produces task-dependent changes in recorded contact. Lid twisting
primarily changes within-pad thumb contact location, whereas egg carton opening
primarily changes index contact occurrence. In matched robot evaluation, revised-interface
demonstrations and the spatial-plus-force input each increase policy success
significantly in the pooled comparisons, and the interface gain appears under both
tactile inputs. Wearable collection systems should
therefore be evaluated not only by how well they map human motion to robot
commands, but also by what physical interactions they make recordable and what
tactile information remains available for learning.

\section*{Acknowledgment}

The authors used Anthropic Claude Code for language editing throughout the
manuscript and to assist with analysis code generation for Sections IV and V. All
analysis code, statistical procedures, numerical results, and interpretations were
reviewed and verified by the authors.

\bibliographystyle{IEEEtran}
\bibliography{refs}

\end{document}